\documentclass[letterpaper]{article} % DO NOT CHANGE THIS
\usepackage{aaai25}  % DO NOT CHANGE THIS
\usepackage{times}  % DO NOT CHANGE THIS
\usepackage{helvet}  % DO NOT CHANGE THIS
\usepackage{courier}  % DO NOT CHANGE THIS
\usepackage[hyphens]{url}  % DO NOT CHANGE THIS
\usepackage{graphicx} % DO NOT CHANGE THIS
\usepackage{natbib}  % DO NOT CHANGE THIS AND DO NOT ADD ANY OPTIONS TO IT
\usepackage{caption} % DO NOT CHANGE THIS AND DO NOT ADD ANY OPTIONS TO IT
\usepackage{algorithm}
\usepackage{algorithmic}
\usepackage{tabularx} 
\usepackage{makecell}
\usepackage{graphicx}
\usepackage{array} 
\usepackage{booktabs} % 在前文導言區添加此行
\usepackage{xcolor} % 在前文導言區添加此行
\usepackage{multirow}
\usepackage{amsmath}
\usepackage{amssymb}
\usepackage[toc,page]{appendix}
\usepackage{newfloat}
\usepackage{listings}
\DeclareCaptionStyle{ruled}{labelfont=normalfont,labelsep=colon,strut=off} % DO NOT CHANGE THIS
\floatstyle{ruled}
\newfloat{listing}{tb}{lst}{}
\floatname{listing}{Listing}
\title{LLM-Based Routing in Mixture of Experts: A Novel Framework for Trading}
\author{
    Kuan-Ming, Liu\textsuperscript{\rm 1}, Ming-Chih,  Lo\textsuperscript{\rm 2}
}
\affiliations{
    \textsuperscript{\rm 1}National Chengchi University, College of Commerce\\
    \textsuperscript{\rm 2}National Yang Ming Chiao Tung University, College of Computer Science\\

}

\usepackage{bibentry}
\begin{document}

\maketitle

\begin{abstract}
Recent advances in deep learning and large language models (LLMs) have facilitated the deployment of the mixture-of-experts (MoE) mechanism in the stock investment domain. 
While these models have demonstrated promising trading performance, they are often unimodal, neglecting the wealth of information available in other modalities, such as textual data. 
Moreover, the traditional neural network-based router selection mechanism fails to consider contextual and real-world nuances, resulting in suboptimal expert selection.
To address these limitations, we propose LLMoE, a novel framework that employs LLMs as the router within the MoE architecture. 
Specifically, we replace the conventional neural network-based router with LLMs, leveraging their extensive world knowledge and reasoning capabilities to select experts based on historical price data and stock news. 
This approach provides a more effective and interpretable selection mechanism.
Our experiments on multimodal real-world stock datasets demonstrate that LLMoE outperforms state-of-the-art MoE models and other deep neural network approaches. 
Additionally, the flexible architecture of LLMoE allows for easy adaptation to various downstream tasks.

\end{abstract}

\section{Introduction}

Traditional trading methods have primarily relied on statistical analysis \cite{DBLP:journals/jsiaml/Kato15} or forecasting models \cite{DBLP:journals/corr/abs-2205-13504}\cite{DBLP:conf/icml/TonerD24}. However, these approaches often struggle to adapt to the complexity and volatility of financial markets, failing to address unseen patterns and dynamic data distributions effectively. In response, deep learning methods have emerged as a promising alternative for quantitative trading \cite{DBLP:conf/kdd/YooSPK21}\cite{DBLP:conf/www/XuL000L21}, offering superior feature learning and insightful market representations. Despite these strengths, deep learning-based algorithms typically rely on a single predictor, leading to performance instability and sensitivity to market fluctuations.

To overcome these limitations, Mixture-of-Experts (MoE) approaches have been introduced \cite{reference1}\cite{ding2024tradexpert}, achieving superior performance and better generalization by leveraging multiple specialized experts. \begin{figure}[ht] \centering \includegraphics[width=\linewidth]{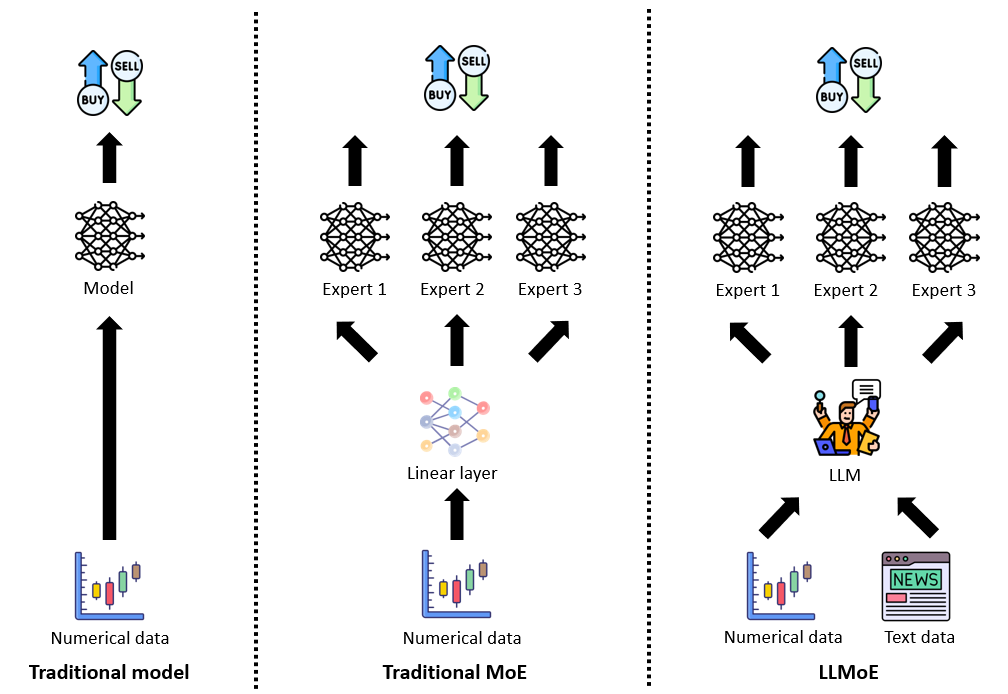} \caption{An illustration comparing traditional single-model approaches, MoE frameworks, and LLMoE. Traditional models use a single predictor with numerical data, MoE adds multiple experts but uses static routing, while LLMoE integrates multimodal data with LLM-driven dynamic routing.} \label{fig:moe-model} \end{figure}

The MoE mechanism in trading algorithms emulates real-world trading rooms, where diverse experts collaborate to tackle specific challenges. While promising, traditional MoE models often suffer from limitations. Routers, typically designed as static neural networks, lack flexibility in financial contexts and are prone to collapse when trained on limited data. Furthermore, current pipelines are predominantly unimodal, relying solely on numerical data and ignoring textual information, such as news, which could provide valuable context and enhance expert selection.

To address these gaps, we propose LLMoE, a novel framework that integrates MoE with advanced language models as routers, as illustrated in Figure~\ref{fig:moe-model}. LLMoE processes historical stock prices and news headlines through an LLM-based router, which provides a comprehensive overview of the current instance. The router dynamically selects the most suitable expert model for stock movement prediction based on the given context. Finally, trading strategies are generated using an "All-in All-out" approach, ensuring robust and informed decision-making. Our experiments demonstrate that LLMoE effectively combines numerical and textual data, enhancing expert selection and achieving superior performance in financial market applications.

\section {Problem Formulation and Methodology}

% \subsection{Problem Formulation}
% Given historical stock feature vectors \( X_{\text{past}} \in \mathbb{R}^{L \times C} \) consisting of \( L \) past time steps and \( C \) features, along with corresponding news information \textit{News} related to the stock, our objective is to predict future stock movements \( Y_{\text{future}} \in \mathbb{R}^T \). In predicting the movement of stock prices, we also aim to develop a corresponding trading strategy based on \( Y_{\text{future}} \), similar to methodologies found in previous studies \cite{DBLP:journals/information/BotunacBM24, reference1}.

\subsection{Problem Formulation}
Given a rolling window of five consecutive descriptive representations:
\[
X_{(t-4:t)} = \{x_{t-4}, x_{t-3}, x_{t-2}, x_{t-1}, x_t\},
\]
where each \( x_i \) is a descriptive string combining numerical features and the corresponding news headline from day \( i \), encapsulating the market conditions of that day. The objective is to predict the stock movement \( Y_{t+1} \in \mathbb{R} \) for the next day.

Additionally, we aim to develop a trading strategy based on \( Y_{t+1} \), leveraging the unified integration of quantitative data and qualitative context for enhanced decision-making. This framework is inspired by methodologies from previous studies~\cite{DBLP:journals/information/BotunacBM24}\cite{reference1}.

\subsection{LLMoE: The LLM-based Router MoE Approach}
\begin{figure}[ht]
    \centering
    \includegraphics[width=\linewidth]{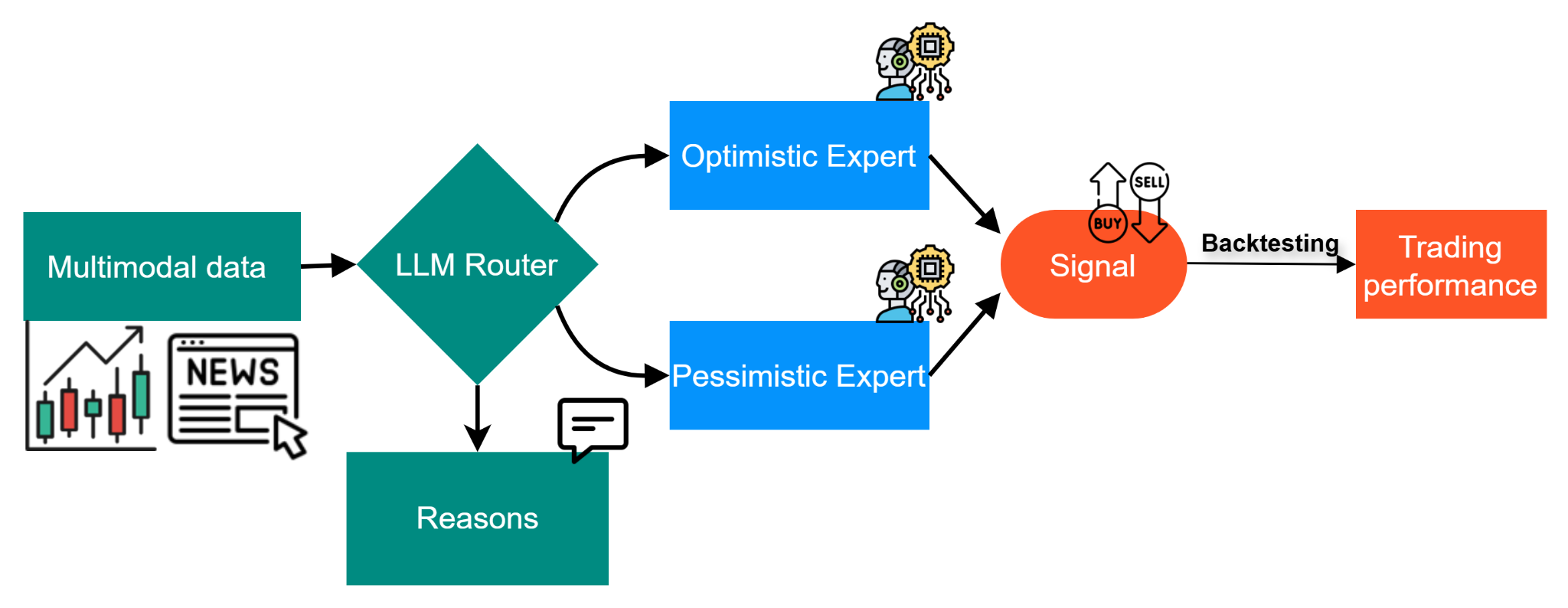} % 圖片寬度可調整
    \caption{Overview of the LLMoE framework, illustrating its three stages: LLM-based router, expert prediction, and trading algorithm generation.}
    \label{fig:llmoe-framework}
\end{figure}

In this work, we propose LLMoE, a novel framework that leverages the power of LLMs to serve as routers within a MoE architecture, thereby providing more efficient expert selection with multimodal data. 
As illustrated in Figure~\ref{fig:llmoe-framework}, our approach consists of three stages: the LLM-based router, expert prediction, and trading algorithm generation.

\subsubsection{LLM-Based Router}
The first stage utilizes an LLM-based router to process and integrate both historical stock price data and relevant news information. 
This router takes advantage of the advanced language comprehension capabilities of LLMs, enabling it to interpret and contextualize multimodal inputs. 
This results in a comprehensive view of the current market conditions. To further enhance expert selection, we categorize the experts based on different contexts, such as positive and negative perspectives. 
Positive experts are trained on instances identified by the router as positive, while negative experts are trained on instances identified as negative. 
This context-based selection mechanism ensures that the most appropriate expert model is chosen to handle each instance. 
By leveraging this strategy, the router can make better-informed decisions, improving the overall expert selection process based on a deeper understanding of the given instances.

% \subsubsection{Expert Prediction}
% The second stage focuses on predictions generated by expert models, which are specifically trained for optimistic and pessimistic market conditions to address diverse market challenges. 
% These expert models leverage data identified by the LLM-based router and perform in-depth analysis using neural network models. 
% % By tailoring predictions to different market scenarios, this context-driven expert selection mechanism not only enhances predictive accuracy but also strengthens the system's decision-making capabilities in dynamic market environments, ultimately resulting in superior and stable backtesting performance.
% Specifically, our expert model is a feedforward neural network (FNN) consisting of fully connected layers designed to capture complex patterns in numerical input features. The model is trained exclusively on numerical data, such as historical price data and related technical indicators, to specialize in different market scenarios. Each expert's architecture is optimized to balance accuracy and computational efficiency, ensuring robust predictions~\cite{reference7}. By tailoring predictions to specific market conditions, this context-driven expert selection mechanism enhances predictive accuracy and strengthens decision-making capabilities, ultimately achieving superior and stable backtesting performance.

\subsubsection{Expert Prediction}
The second stage focuses on predictions generated by expert models trained for optimistic and pessimistic market conditions. 
These feedforward neural networks (FNNs) leverage data identified by the LLM-based router, analyzing numerical inputs such as prices indicators to address specific market scenarios. 
Optimized for accuracy and efficiency, these experts enhance predictive accuracy and decision-making, achieving stable and robust backtesting performance~\cite{reference7}.
Temporal Graph Networks for Graph Anomaly Detection in Financial Networks

\subsubsection{Trading Algorithm Generation}
In the final stage, the selected expert's predictions are utilized to generate robust trading strategies. 
We employ an "All-in All-out" strategy, where all available cash is invested when the expert predicts positive price movements, and all holdings are liquidated when the expert predicts negative price movements. 
This strategy aims to maximize returns by dynamically adjusting investment positions based on the expert model's output.

\section{Experiment}

\subsection{Experiment Setup}

\subsubsection{1. Datasets}

To evaluate LLMoE, we use two US market datasets spanning a decade (2006--2016), combining stock price data and news headlines for multimodal analysis. 
The MSFT dataset, with significant missing news days, presents a challenging test for handling incomplete data, while the AAPL dataset provides a more complete environment. 
These datasets comprehensively evaluate LLMoE’s ability to integrate multimodal data under varying conditions. Further details on dataset composition and splits are provided in Appendix - Experiment Setup Details. 
% ~\ref{appendix-datasets}.

\subsubsection{2. Features}

To effectively capture market dynamics, we engineered a set of features encompassing price ratios, daily price changes, and rolling deviations calculated from moving averages of various lengths. These features are designed to capture both short-term and long-term market trends, providing comprehensive insights into market behavior.Further details on feature calculations and formulas are provided in Appendix - Experiment Setup Details. 
% ~\ref{sec:appendix-features}.

% \subsubsection{2. Features\\}
% To effectively capture market dynamics, we engineered a range of features that encompass price ratios ($z_{open}$, $z_{high}$, $z_{low}$), daily price changes ($z_{close}$, $z_{adjclose}$), and rolling deviations ($zd_{n}$) calculated from $n$-day moving averages, where $n \in \{5, 10, 15, 20, 25, 30\}$. These features are designed to capture both short-term and long-term market trends. For instance, $zd_{20}$ quantifies the deviation of the adjusted closing price from its 20-day moving average, providing insights into medium-term market movements.

% \begin{table}[h!]
%     \centering
%     \renewcommand{\arraystretch}{1.3}
%     \caption{Calculation Formulas for Stock Market Features}
%     \label{tab:feature_formulas}
%     \begin{tabular}{|l|l|}
%         \hline
%         \textbf{Features} & \textbf{Calculation Formula} \\
%         \hline
%         $z_{open}, z_{high}, z_{low}$ & $z_{open} = \frac{open_t}{close_t} - 1$ \\
%         \hline
%         $z_{close}, z_{adjclose}$ & $z_{close} = \frac{close_t}{close_{t-1}} - 1$ \\
%         \hline
%         \makecell[l]{$zd_5, zd_{10}, zd_{15},$\\$zd_{20}, zd_{25}, zd_{30}$} & $zd_5 = \frac{\sum_{i=0}^{4} adj\_close_{t-i} / 5}{adj\_close_t} - 1$ \\
%         \hline
%     \end{tabular}
% \end{table}

\subsubsection{3. Baseline Models}

To evaluate the effectiveness of the proposed LLMoE framework, we compare it against several baseline models, including gradient boosting, neural networks, and traditional Mixture of Experts (MoE) models. These models provide a diverse set of benchmarks to evaluate LLMoE's performance. Detailed descriptions of the baseline models are provided in Appendix - Experiment Setup Details. 
% ~\ref{sec:appendix-baselines}.

% \subsubsection{3. Baseline Models\\\\}

% To evaluate the effectiveness of the proposed LLMoE framework, we compare it against the following baseline models:

% \begin{itemize}
%     \item \textbf{LightGBM (LGB)}: A gradient boosting model widely used for structured data, offering strong performance in various classification tasks.
    
%     \item \textbf{Multi-Layer Perceptron (MLP)}: A fully connected neural network, designed to model complex relationships in tabular data.
    
%     \item \textbf{Long Short-Term Memory (LSTM)}: A recurrent neural network capable of capturing temporal dependencies in sequential data, particularly suited for stock price trends.
    
%     \item \textbf{Dynamic Neural Network Ensemble (DNNE)}: An ensemble model that combines multiple neural networks trained on bootstrap samples to enhance prediction diversity and robustness.
    
%     \item \textbf{Mixture of Experts (MoE)}: Traditional MoE models employing static routing mechanisms, evaluated with ensemble sizes of 2 and 10 experts.
% \end{itemize}

% These baseline models provide a comprehensive comparison, showcasing their strengths and limitations when applied to financial classification tasks.
\subsubsection{4. Evaluation Metrics}

To evaluate the performance of our trading models, we employ seven commonly used financial metrics: Total Return (TR), Annualized Volatility (VOL), Sharpe Ratio (SR), Sortino Ratio (SoR), Maximum Drawdown (MDD), Calmar Ratio (CR), and Downside Deviation (DD). These metrics collectively measure the models' ability to balance returns and risks under different market conditions.Detailed definitions and formulas for these metrics are provided in Appendix - Experiment Setup Details. 

\subsubsection{5. Other Experimental Details}

Hyperparameters for baseline models, including learning rates and hidden layer sizes, were optimized using random search across three trials. For LLMoE, grid search focused on batch size, hidden layer size, and learning rate. All models used a consistent 5-day look-back window for fair comparisons. To ensure robust performance evaluation, experiments were repeated with ten random seeds, and confidence intervals were calculated from the standard deviation of metrics across these trials.

% Hyperparameters for baseline models, including learning rates and hidden layer sizes, were optimized using random search across three trials. For LLMoE, grid search focused on key neural network hyperparameters such as batch size, hidden layer size, and learning rate. All models used a consistent 5-day look-back window to capture short-term market trends and ensure fair comparisons.

% \subsubsection{5. Other Experimental Details}
% \begin{itemize}
%     \item \textbf{Hyperparameter Tuning.}: Random search optimized baseline hyperparameters like learning rates and hidden layers in three trials, while grid search fine-tuned LLMoE configurations, including routing and expert count.
%     % Random search was applied to optimize hyperparameters like learning rates and hidden layer sizes for baseline models, with each model undergoing three trials to ensure a balance between efficiency and performance. For the LLMoE framework, grid search was used to fine-tune its unique configurations, such as the routing mechanism and number of experts.
%     \item \textbf{Look-Back Window.}: A 5-day look-back window was adopted for input feature construction across all models. This setup captures short-term market trends and ensures consistency in temporal data representation for fair comparisons.

% \end{itemize}

\subsection{Implementation and Evaluation}
\subsubsection{1. Router\\}

We use Llama3.2 as our router, which serves as a critical component in the LLMoE framework, to classify the market outlook based on a five-day rolling window of features and descriptions, facilitating the integration of numerical and textual data for decision-making.

\paragraph{Input Features and Descriptions.}  
The input to the router consists of a rolling window of five consecutive data points:
\[
X_{(t-4:t)} = \{x_{t-4}, x_{t-3}, x_{t-2}, x_{t-1}, x_t\}
\]
% Each \( x_i \) represents a combination of numerical features and the corresponding news headline from day \( i \).
Each \( x_i \) combines numerical features with the corresponding news headline from day \( i \), forming a single descriptive string that encapsulates the market conditions of that day. 
This representation integrates quantitative data and qualitative context into a unified format for analysis.

\paragraph{Router Outputs}  
The LLM-based router provides two key outputs to facilitate classification and interpretability:

\begin{itemize}
    \item \textbf{Classification:}  
    The router evaluates the rolling window of numerical and textual data, assigning a label as either \textbf{Optimistic} or \textbf{Pessimistic}, reflecting the predicted market sentiment:
    \[
    Router(X_{(t-4:t)}) \to \{Optimistic, Pessimistic\}
    \]
    The label \( y_t \) is selected based on the highest likelihood:
    \[
    y_t = \arg\max_{Optimistic, Pessimistic}(\text{LLM Prediction}(X_{(t-4:t)}))
    \]

    \item \textbf{Reasoning:}  
    To improve interpretability, the router generates a natural language explanation \( R \), offering insights into the factors influencing its classification:
    \[
    R = \text{LLM Reason Output}
    \]
    This reasoning enhances transparency without directly affecting subsequent expert models.
\end{itemize}

\subsection{2. Expert Models}

The expert models for optimistic and pessimistic market conditions share a unified architecture, designed to process numerical features and predict the direction of the next day's stock price movement. The architecture begins with an input layer that processes \( n = 55 \) numerical features derived from daily market data through feature engineering. These features are organized into a rolling window structure:
\[
X_{(t-4:t)} = \{x_{t-4}, x_{t-3}, x_{t-2}, x_{t-1}, x_t\},
\]
where each \( x_i \) represents 11 numerical attributes, encompassing daily price metrics and rolling deviations:
\begin{equation*}
\begin{split}
x_i = \{&z_{\text{open}}, z_{\text{high}}, z_{\text{low}}, z_{\text{close}}, z_{\text{adjclose}},\\
&zd_5, zd_{10}, zd_{15}, zd_{20}, zd_{25}, zd_{30}\}.
\end{split}
\end{equation*}

This input representation ensures that the model captures both short-term fluctuations and long-term trends, enabling it to generate accurate predictions. Further details on the hidden layers and output layer configuration are provided in Appendix - Expert Model Architecture. 

\subsection{3. Experiment Results}

\begin{table*}[t] % t! 強制置頂
    \centering
    \setlength{\tabcolsep}{6pt}
    \renewcommand{\arraystretch}{1.3}
    \caption{Performance Comparison of Different Models}
    \label{table-performance} % 添加標籤
    \resizebox{\linewidth}{!}{
        \begin{tabular}{|c|l|l||c|c|c|c|c|c|c|} % 加入雙直線分隔符號
            \hline
            \textbf{Market} & \textbf{Type} & \textbf{Model} & \textbf{TR} & \textbf{SR} & \textbf{CR} & \textbf{SoR} & \textbf{VOL} & \textbf{DD} & \textbf{MDD} \\ \hline
            \multirow{7}{*}{MSFT} 
            
            & ENS                                & DNNE                                & 9.48±8.19                        & 0.51±0.36                        & 0.78±0.71                        & 0.66±0.43                         & 0.27±0.04                         & 0.39±0.08                        & 16.69±4.52                        \\ 
            & RNN                                & LSTM                                & 27.93±8.20                       & \underline{1.39±0.31}                  & 2.26±0.87                        & \underline{1.84±0.45}                   & 0.38±0.03                         & 0.42±0.03                        & 14.39±2.60                        \\
            & BDT                                & LGB                                 & 30.63±7.58                       & 1.05±0.21                        & 2.46±0.72                        & 1.14±0.25                         & \textbf{0.25±0.01}                & \textbf{0.33±0.03}               & \underline{13.86±2.44}                  \\ 
            & NRNN                               & MLP                                 & \underline{33.92±7.59}                 & 1.21±0.22                        & \underline{2.54±0.81}                  & 1.33±0.35                         & 0.27±0.02                         & 0.36±0.04                        & 14.35±2.12                        \\ 
            & MoE                                & MoE\_10                             & 10.84±8.85                       & 0.45±0.35                        & 0.78±0.62                        & 0.5±0.32                          & 0.27±0.02                         & 0.39±0.04                        & 17.69±2.71                        \\ 
            & MoE                                & MoE\_2                              & 22.18±17.13                      & 0.78±0.60                        & 1.86±1.65                        & 0.96±0.78                         & \underline{0.26±0.01}                   & 0.38±0.04                        & 17.82±3.57                        \\ 
            & \textbf{MoE}                                & \textbf{LLMoE}                               & \textbf{65.44±11.10}             & \textbf{2.14±0.29}               & \textbf{5.91±1.12}               & \textbf{2.24±0.37}                & 0.26±0.02                         & \underline{0.34±0.03}                  & \textbf{11.32±1.09}               \\ \hline

            \multirow{7}{*}{AAPL} 
            & ENS                                & DNNE                                & -3.66±4.87                       & -0.09±0.30                       & -0.11±0.19                       & -0.01±0.35                        & 0.62±0.14                         & 0.77±0.12                        & 25.34±3.29                        \\ 
            & RNN                                & LSTM                                & 18.04±9.14                       & 0.85±0.43                        & 1.3±0.84                         & 1.17±0.66                         & 0.34±0.02                         & 0.42±0.06                        & \underline{17.84±2.96}                  \\
            & BDT                                & LGBM                                & 8.65±7.22                        & 0.39±0.28                        & 0.4±0.36                         & 0.63±0.39                         & \textbf{0.28±0.03}                & \textbf{0.3±0.02}                & 26.14±2.91                        \\ 
            & NRNN                               & MLP                                 & \underline{26.16±6.35}                 & \underline{0.99±0.21}                  & \underline{1.76±0.69}                  & \underline{1.31±0.33}                   & 0.3±0.02                          & \underline{0.33±0.02}                  & \textbf{17.19±2.68}               \\
            
            & MoE                                & MoE\_10                             & 8.77±10.41                       & 0.41±0.44                        & 0.5±0.51                         & 0.61±0.54                         & 0.43±0.08                         & 0.55±0.11                        & 23.44±2.95                        \\
            & MoE                                & MoE\_2                              & 6.73±8.82                        & 0.32±0.37                        & 0.47±0.48                        & 0.48±0.42                         & 0.35±0.04                         & 0.44±0.05                        & 24.83±5.17                        \\ 
            & \textbf{MoE}                                & \textbf{LLMoE}                               & \textbf{31.43±11.46}             & \textbf{1.17±0.33}               & \textbf{2.12±1.10}               & \textbf{1.37±0.48}                & \underline{0.29±0.04}                   & 0.36±0.04                        & 18.21±4.23                        \\\hline
        \end{tabular}
    }
    \label{tab:performance_comparison}
    % Additional explanation for the table
    \captionsetup{justification=justified,singlelinecheck=false,font=footnotesize}
    \caption*{
        \textbf{Note:} For each metric, the best-performing model is highlighted in \textbf{bold}, while the second-best is underlined. The results indicate that LLMoE consistently outperforms other baseline models across most metrics, demonstrating its superiority in both return and risk-adjusted performance. Particularly, LLMoE achieves the highest Total Return (TR), Sharpe Ratio (SR), and Calmar Ratio (CR) on both MSFT and AAPL datasets, reflecting its robust and adaptive approach to multimodal data integration.
}
\end{table*}

\subsubsection{The Router’s Human-Like Reasoning\\}

The router in the LLMoE framework demonstrates human-like reasoning by integrating numerical data and textual information. 
For example, in a specific instance, "\textit{despite concerns about Apple's growth highlighted in news headlines}", the router identified "\textit{consistent increases in prices and volume}", which suggested a "\textit{cautiously optimistic outlook}". 
This reasoning showcases the router’s ability to weigh conflicting signals—optimistic numerical trends against mixed textual sentiment—allowing it to produce well-balanced and context-aware predictions.

\subsubsection{Outstanding Performance of LLMoE\\}
Our LLMoE model significantly outperformed other baseline models in key metrics, including Total Return (TR), Sharpe Ratio (SR), and Calmar Ratio (CR), demonstrating superior performance in balancing returns and risks, as shown in Table~\ref{table-performance}. 
This highlights the efficiency and accuracy of using LLMs as routers to integrate numerical and textual data.

\subsubsection{Comparison Between 2-Expert MoE and LLMoE\\}
LLMoE demonstrated clear superiority over the 2-expert MoE model by leveraging LLMs as intelligent routers. 
Unlike the 2-expert MoE, which relies on static routing, LLMoE dynamically integrates multimodal data, enabling more effective allocation of expert resources. 
This resulted in significantly better performance in risk-adjusted return metrics, such as the Sharpe Ratio (SR) and Calmar Ratio (CR), as well as improved risk management with a lower Maximum Drawdown (MDD).

\section{Conclusion}
In this paper, we present LLMoE, a novel framework that integrates a pre-trained Large Language Model (LLM) as a router within a Mixture of Experts (MoE) architecture. By dynamically combining numerical stock features with textual news data, LLMoE bridges the gap between quantitative and qualitative analysis, enabling accurate and interpretable predictions for financial markets. This dynamic and context-aware routing mechanism surpasses traditional MoE systems' static limitations, enhancing adaptability to volatile market conditions. Our experimental results demonstrate LLMoE’s superior performance, achieving over 25\% improvements in key risk-adjusted return metrics like the Sharpe Ratio and Total Return, establishing it as a state-of-the-art tool for intelligent trading strategies.

\bibliographystyle{aaai25}
\bibliography{aaai25}

\begin{thebibliography}{14}
\providecommand{\natexlab}[1]{#1}

\bibitem[{Botunac, Bosna, and Matetic(2024)}]{DBLP:journals/information/BotunacBM24}
Botunac, I.; Bosna, J.; and Matetic, M. 2024.
\newblock Optimization of Traditional Stock Market Strategies Using the {LSTM} Hybrid Approach.
\newblock \emph{Inf.}, 15(3): 136.

\bibitem[{Ding, Shi, and Liu(2024)}]{ding2024tradexpert}
Ding, Q.; Shi, H.; and Liu, B. 2024.
\newblock TradExpert: Revolutionizing Trading with Mixture of Expert LLMs.
\newblock \emph{arXiv preprint arXiv:2411.00782}.

\bibitem[{Hu et~al.(2018)Hu, Liu, Bian, Liu, and Liu}]{reference4}
Hu, Z.; Liu, W.; Bian, J.; Liu, X.; and Liu, T.-Y. 2018.
\newblock A Deep Learning Framework for News-oriented Stock Trend Prediction.
\newblock \emph{Proceedings of the 11th ACM International Conference on Web Search and Data Mining}, 297--305.

\bibitem[{Kato(2015)}]{DBLP:journals/jsiaml/Kato15}
Kato, T. 2015.
\newblock {VWAP} execution as an optimal strategy.
\newblock \emph{{JSIAM} Lett.}, 7: 33--36.

\bibitem[{Kou et~al.(2024)Kou, Yu, Peng, and Chen}]{reference7}
Kou, Z.; Yu, H.; Peng, J.; and Chen, L. 2024.
\newblock Automate Strategy Finding with {LLM} in Quant investment.
\newblock \emph{CoRR}, abs/2409.06289.

\bibitem[{Li and Xu(2023)}]{reference6}
Li, K.; and Xu, J. 2023.
\newblock An Attention-Based Multi-Gate Mixture-of-Experts Model for Quantitative Stock Selection.
\newblock \emph{International Journal of Trade, Economics and Finance}, 14(3): 165--173.

\bibitem[{Sawhney et~al.(2020)Sawhney, Agarwal, Wadhwa, and Shah}]{reference3}
Sawhney, R.; Agarwal, S.; Wadhwa, A.; and Shah, R.~R. 2020.
\newblock Deep Attentive Learning for Stock Movement Prediction From Social Media Text and Company Correlations.
\newblock 8415--8426.

\bibitem[{Sun et~al.(2023)Sun, Wang, Xue, Lou, and An}]{reference1}
Sun, S.; Wang, X.; Xue, W.; Lou, X.; and An, B. 2023.
\newblock Mastering Stock Markets with Efficient Mixture of Diversified Trading Experts.
\newblock \emph{Proceedings of the 29th ACM SIGKDD Conference on Knowledge Discovery and Data Mining}, 2109--2119.

\bibitem[{Toner and Darlow(2024)}]{DBLP:conf/icml/TonerD24}
Toner, W.; and Darlow, L.~N. 2024.
\newblock An Analysis of Linear Time Series Forecasting Models.
\newblock In \emph{Forty-first International Conference on Machine Learning, {ICML} 2024, Vienna, Austria, July 21-27, 2024}. OpenReview.net.

\bibitem[{Xu et~al.(2021)Xu, Liu, Xu, Bian, Yin, and Liu}]{DBLP:conf/www/XuL000L21}
Xu, W.; Liu, W.; Xu, C.; Bian, J.; Yin, J.; and Liu, T. 2021.
\newblock {REST:} Relational Event-driven Stock Trend Forecasting.
\newblock In Leskovec, J.; Grobelnik, M.; Najork, M.; Tang, J.; and Zia, L., eds., \emph{{WWW} '21: The Web Conference 2021, Virtual Event / Ljubljana, Slovenia, April 19-23, 2021}, 1--10. {ACM} / {IW3C2}.

\bibitem[{Yoo et~al.(2021{\natexlab{a}})Yoo, Soun, chan Park, and Kang}]{reference2}
Yoo, J.; Soun, Y.; chan Park, Y.; and Kang, U. 2021{\natexlab{a}}.
\newblock Accurate Multivariate Stock Movement Prediction via Data-Axis Transformer with Multi-Level Contexts.
\newblock \emph{Proceedings of the 27th ACM SIGKDD Conference on Knowledge Discovery and Data Mining}, 313--323.

\bibitem[{Yoo et~al.(2021{\natexlab{b}})Yoo, Soun, Park, and Kang}]{DBLP:conf/kdd/YooSPK21}
Yoo, J.; Soun, Y.; Park, Y.; and Kang, U. 2021{\natexlab{b}}.
\newblock Accurate Multivariate Stock Movement Prediction via Data-Axis Transformer with Multi-Level Contexts.
\newblock In Zhu, F.; Ooi, B.~C.; and Miao, C., eds., \emph{{KDD} '21: The 27th {ACM} {SIGKDD} Conference on Knowledge Discovery and Data Mining, Virtual Event, Singapore, August 14-18, 2021}, 2037--2045. {ACM}.

\bibitem[{Yu et~al.(2024)Yu, Wu, Wang, and Weng}]{reference5}
Yu, Z.; Wu, Y.; Wang, G.; and Weng, H. 2024.
\newblock MIGA: Mixture-of-Experts with Group Aggregation for Stock Market Prediction.
\newblock \emph{arXiv preprint arXiv:2410.02241}.

\bibitem[{Zeng et~al.(2022)Zeng, Chen, Zhang, and Xu}]{DBLP:journals/corr/abs-2205-13504}
Zeng, A.; Chen, M.; Zhang, L.; and Xu, Q. 2022.
\newblock Are Transformers Effective for Time Series Forecasting?
\newblock \emph{CoRR}, abs/2205.13504.

\end{thebibliography}

\appendix
\section{Appendix}
\subsection{Related Work}
\subsubsection{Financial Prediction with Deep Learning\\}

Deep learning methods for financial prediction leverage RNNs, such as LSTM and GRU, to capture temporal patterns, and NRNNs, like transformers and graph-based models, to analyze inter-stock relationships and market dynamics~\cite{reference2}\cite{reference3}\cite{reference4}. Additionally, alternative data sources, such as tweets and news, improve predictions~\cite{reference4}\cite{reference3}. However, these approaches often require high computational resources and lack robustness in volatile markets. Lightweight frameworks like \textit{AlphaMix} address these challenges by using simple MLP backbones, achieving comparable predictive performance while reducing computational costs~\cite{reference1}. By integrating diverse data sources, including historical prices and alternative data, \textit{AlphaMix} enhances robustness in highly stochastic markets~\cite{reference1}\cite{reference4}.

% \subsection{Ensemble Learning}
% Ensemble learning methods, which combine the outputs of multiple models to improve generalization, have evolved significantly. Techniques like Snapshot Ensemble leverage cyclic learning rates to explore multiple local minima~\cite{reference1, reference2}, while BatchEns optimizes computation using Hadamard products~\cite{reference1}. Model Soup, which averages neural network weights, offers excellent performance without additional inference costs~\cite{reference1}.

% Despite these advancements, ensemble methods for financial prediction primarily rely on traditional techniques such as bagging and stacking, which scale poorly in dynamic environments~\cite{reference3}. AlphaMix addresses this limitation by customizing advanced ensemble methods to efficiently capture fleeting trading opportunities while minimizing computational overhead~\cite{reference1, reference2}.

\subsubsection{Mixture of Experts\\}
The mixture-of-experts (MoE) framework, widely used in computer vision and natural language processing for scalability and multi-task learning~\cite{reference5}\cite{reference6}, remains underexplored in quantitative finance. Although \textit{AlphaMix} improves accuracy and efficiency with a three-stage design~\cite{reference1}, it relies on manual routing mechanisms, limiting expert specialization and multimodal integration to structured data while excluding unstructured sources like news. Moreover, its routing lacks interpretability, reducing transparency in decision-making. To address these limitations, we propose an MoE framework utilizing LLMs as adaptive routers. By dynamically selecting experts based on multimodal inputs such as historical prices and alternative data, our approach enhances interpretability, adaptability, and robustness in volatile financial markets.
% The mixture-of-experts (MoE) framework has been widely applied in fields like computer vision and natural language processing, with advancements such as sparse MoE layers for scalability~\cite{reference5} and multi-gate frameworks for multi-task learning~\cite{reference6}.

% In quantitative finance, MoE frameworks are underexplored. 
% While AlphaMix improves accuracy and efficiency with a three-stage design~\cite{reference1}, it lacks clear expert specialization due to its reliance on manual routing mechanisms. 
% Its multimodal integration is limited to structured data, excluding rich unstructured sources like news. 
% Additionally, its routing lacks interpretability, making decision processes less transparent for practical use.

% To address these gaps, we propose an MoE framework with LLMs as adaptive routers. 
% By defining expert specializations based on multimodal inputs like historical prices and alternative data, and dynamically selecting experts through LLM reasoning, our approach enhances interpretability, adaptability, and robustness in volatile financial markets.

\subsection{Experiment Setup Details}

\subsubsection{Datasets\\}\label{appendix-datasets}

The datasets include numerical stock price data and textual news information, offering a robust environment for multimodal integration. 
For these two datasets:
\begin{itemize}
    \item \textbf{MSFT Dataset}: Includes daily stock prices and news headlines. Notably, 1,176 of 2,503 trading days lack news, making it a challenging test for LLMoE's handling of incomplete multimodal data.
    \item \textbf{AAPL Dataset}: Provides 2,482 trading days, with only 194 missing news entries. This more complete dataset complements MSFT by showcasing LLMoE's adaptability.
\end{itemize}

\paragraph{Dataset Splits}  
The training period spans the first 80\% of trading days (\textbf{2006-12-07} to \textbf{2014-12-02}), while the testing period covers the remaining 20\% (\textbf{2014-12-03} to \textbf{2016-11-29}). These splits test model robustness under varying levels of news coverage and provide comprehensive evaluation conditions.

\begin{table}[h!]
    \centering
    \setlength{\tabcolsep}{4pt} % 調整單元格內間距
    \renewcommand{\arraystretch}{1.2} % 調整行高
    \caption{Dataset Details for MSFT and AAPL}
    \label{tab:dataset_details}
    \resizebox{\linewidth}{!}{ % 限制表格寬度在頁面範圍內
        \begin{tabular}{|l|c|c|c|c|}
        \hline
        \textbf{Dataset} & \textbf{Days} & \textbf{No News Days} & \textbf{From} & \textbf{To} \\ \hline
        MSFT             & 2,503         & 1,176                 & 06/12/01      & 16/11/30    \\ \hline
        AAPL             & 2,482         & 194                   & 06/12/01      & 16/11/30    \\ \hline
        \end{tabular}
    }
\end{table}

\subsubsection{Feature Engineering\\}\label{sec:appendix-features}

To effectively capture market dynamics, we engineered a range of features that encompass:
\begin{itemize}
    \item \textbf{Price Ratios} ($z_{open}$, $z_{high}$, $z_{low}$): These quantify the ratio between opening, high, and low prices relative to the closing price.
    \item \textbf{Daily Price Changes} ($z_{close}$, $z_{adjclose}$): These capture the daily percentage change in closing and adjusted closing prices.
    \item \textbf{Rolling Deviations} ($zd_{n}$): Calculated from $n$-day moving averages, where $n \in \{5, 10, 15, 20, 25, 30\}$, these features quantify deviations over varying time horizons to provide insights into market trends.
\end{itemize}

For example, $zd_{20}$ quantifies the deviation of the adjusted closing price from its 20-day moving average, offering insights into medium-term market movements.

\begin{table}[h!]
    \centering
    \renewcommand{\arraystretch}{1.3}
    \caption{Calculation Formulas for Stock Market Features}
    \label{tab:feature_formulas}
    \begin{tabular}{|l|l|}
        \hline
        \textbf{Features} & \textbf{Calculation Formula} \\
        \hline
        $z_{open}, z_{high}, z_{low}$ & $z_{open} = \frac{open_t}{close_t} - 1$ \\
        \hline
        $z_{close}, z_{adjclose}$ & $z_{close} = \frac{close_t}{close_{t-1}} - 1$ \\
        \hline
        \makecell[l]{$zd_5, zd_{10}, zd_{15},$\\$zd_{20}, zd_{25}, zd_{30}$} & $zd_5 = \frac{\sum_{i=0}^{4} adj\_close_{t-i} / 5}{adj\_close_t} - 1$ \\
        \hline
    \end{tabular}
\end{table}
\subsubsection{Baseline models\\}\label{sec:appendix-baselines}

To evaluate the effectiveness of the proposed LLMoE framework, we compare it against the following baseline models:

\begin{itemize}
    \item \textbf{LightGBM (LGB)}: A gradient boosting model widely used for structured data, offering strong performance in various classification tasks. It is included to assess how tree-based models compare to neural networks in handling tabular financial data.

    \item \textbf{Multi-Layer Perceptron (MLP)}: A fully connected neural network, designed to model complex relationships in tabular data. This serves as a baseline for deep learning on financial features.

    \item \textbf{Long Short-Term Memory (LSTM)}: A recurrent neural network capable of capturing temporal dependencies in sequential data, particularly suited for stock price trends. It provides a benchmark for time-series models.

    \item \textbf{Dynamic Neural Network Ensemble (DNNE)}: An ensemble model that combines multiple neural networks trained on bootstrap samples to enhance prediction diversity and robustness. It serves as a comparison point for ensemble methods.

    \item \textbf{Mixture of Experts (MoE)}: Traditional MoE models employing static routing mechanisms, evaluated with ensemble sizes of 2 and 10 experts. These models provide insights into the advantages of dynamic routing introduced by LLMoE.
\end{itemize}

These baseline models offer a comprehensive set of benchmarks, showcasing their respective strengths and limitations in financial classification tasks.

\subsubsection{Evaluation Metrics\\}\label{sec:appendix-metrics}

To comprehensively evaluate the performance of trading models, we employ the following financial metrics:

\begin{itemize}
    \item \textbf{Total Return (TR)}: The percentage change in the portfolio value over the entire trading period, calculated as:
    \[
    TR = \frac{\text{Final Portfolio Value} - \text{Initial Portfolio Value}}{\text{Initial Portfolio Value}} \times 100
    \]

    \item \textbf{Annualized Volatility (VOL)}: A measure of the portfolio's risk, representing the standard deviation of daily returns scaled to an annualized basis:
    \[
    VOL = \text{Standard Deviation of Daily Returns} \times \sqrt{252}
    \]

    \item \textbf{Sharpe Ratio (SR)}: Evaluates the risk-adjusted return of the portfolio, defined as the ratio of the mean excess return (over the risk-free rate) to its standard deviation:
    \[
    SR = \frac{\text{Mean Excess Return}}{\text{Standard Deviation of Returns}}
    \]

    \item \textbf{Sortino Ratio (SoR)}: A variation of the Sharpe Ratio focusing only on downside risk, calculated as:
    \[
    SoR = \frac{\text{Mean Excess Return}}{\text{Downside Deviation}}
    \]

    \item \textbf{Maximum Drawdown (MDD)}: The largest peak-to-trough decline in portfolio value during the trading period, expressed as a percentage:
    \[
    MDD = \max\left(\frac{\text{Peak Value} - \text{Trough Value}}{\text{Peak Value}}\right) \times 100
    \]

    \item \textbf{Calmar Ratio (CR)}: Measures risk-adjusted return by considering the ratio of total return to maximum drawdown:
    \[
    CR = \frac{\text{Total Return}}{\text{Maximum Drawdown}}
    \]

    \item \textbf{Downside Deviation (DD)}: Captures the standard deviation of negative returns, emphasizing periods of underperformance:
    \[
    DD = \sqrt{\frac{\sum (\text{Negative Returns})^2}{\text{Number of Observations}}}
    \]
\end{itemize}

\subsection{Expert Model Architecture}\label{sec:appendix-expert}

The expert models for optimistic and pessimistic market conditions share a common architecture, consisting of the following components:

\begin{itemize}
    \item \textbf{Hidden Layers:}  
    The model includes three fully connected dense layers to capture complex patterns in the input data:
    \begin{itemize}
        \item \textbf{Layer 1}: 128 neurons with ReLU activation.
        \item \textbf{Layer 2}: 64 neurons with ReLU activation and a dropout rate of 0.3.
        \item \textbf{Layer 3}: 32 neurons with ReLU activation and a dropout rate of 0.2.
    \end{itemize}
    These layers are optimized to balance model capacity and regularization, ensuring the ability to generalize to unseen data.

    \item \textbf{Output Layer:}  
    The final output layer consists of a single neuron with a Sigmoid activation function. It outputs a binary classification indicating the direction of the next day's stock price movement (increase or decrease). The predicted direction supports trading strategy generation, offering actionable insights into market trends.
\end{itemize}

\end{document}